# Rigid chain in parallel kinematic positioning system

M Kubrikov[1], I Pikalov[1,2] and M Saramud[1,2]

[1] Reshetnev Siberian State University of Science & Technology, 31 Krasnoyarsky Rabochy Av., Krasnoyarsk, 660037, Russia
[2] Siberian Federal University, 79 Svobodny Av., Krasnoyarsk, 660041, Russia

E-mail: gidroponika@mail.ru, yapibest@mail.ru, msaramud@gmail.com

**Abstract**. The article presents an analysis of the trends in the development of kinematic structures of modern machine-building technological equipment. The prospects of using machines with parallel kinematics in processing, measuring and handling equipment, their advantages and disadvantages are demonstrated. It is shown that it is inexpedient to use ball screw drives in machines with parallel kinematics, performing tasks of low accuracy, but with displacements of more than 3000 mm. The rigid chain system of the Serapid firm and the possibility of its use in machines with parallel kinematics are considered. A schematic solution of a three-coordinate manipulation robot based on parallel kinematics with drive mechanisms on rigid chains is proposed.

## 1. Introduction
The development of Industry 4.0 and cloud production requires the introduction of new methods and approaches to production processes. In cloud production, manual labor is minimized [1, 2], technological processes using manual labor are replaced by fully robotic ones. Often, production is located within a single production facility, which includes warehouse space, processing equipment, assembly and locksmith areas, and more [3]. In a generalized form, the production process involves the receipt of a workpiece, tool and technological equipment in the warehouse, their movement and installation on the machine, subsequent processing and transportation of the finished part to the assembly and locksmith area, from where the finished product is transferred to the control zone of the finished product or to the warehouse [4]. In traditional manufacturing, manual labor is used to organize these operations. The article discusses a way to minimize manual labor within a multifunctional production facility. The proposed method describes a positioning system with parallel kinematics based on drives with a rigid chain, in order to ensure movement of the product throughout the entire technological cycle, that is, not only when moving from section to section, but also when positioning the workpiece on the machine.

## 2. The relevance of research
In modern mechanical engineering, a large range of technological equipment is used: metalworking machines, automatic welding machines, coordinate measuring machines, manipulation robots [5]. In most cases, they have a traditional (sequential) kinematic structure, when one drive is responsible for movement along each of the Cartesian axes (or rotation around an axis), respectively, the performance characteristics (speed, acceleration, stiffness and positioning accuracy) along each of the axes are also determined one drive. The forces acting on the bearing elements of the machine, as a rule, lead to bending stresses, which can be a source of low rigidity (and, consequently, accuracy) of the executive links of the machine. To ensure rigidity and positioning accuracy, profile guides are mainly used, which require high accuracy of their installation when assembling equipment.

Equipment with parallel kinematics in which movement along one axis (or rotation) is provided not by one drive mechanism, but by the joint operation of several (usually linear) drives [8] is becoming more and more widespread [6, 7]. Such machines have a complex spatial structure in which the bearing elements work under the action of tensile and compressive stresses, which significantly increases the rigidity and accuracy of movements. In addition, due to the reduction in the mass of the moving elements, as well as the use of several drives, it is possible to increase the speed and acceleration of movements. The support of the actuator in space is carried out through the use of rigid rods.

Ball screw drives [9] with servo motors are the most widely used drive mechanisms. This is due to its many advantages over other types of transmissions, such as high efficiency (up to 90%), high service life and rigidity, ease of repair and maintenance, high positioning accuracy, and is able to provide sufficient acceleration and displacement speeds. However, these mechanisms also have disadvantages: firstly, the length is limited to 3000 mm; secondly, when realizing machines with parallel kinematics, about half of the lead screw length will go beyond the overall dimensions of the working area (thereby increasing the space occupied by the machine).

In addition to high-precision machines with parallel kinematics (metalworking machines and coordinate measuring machines), there is a separate group of technological machines that do not require high precision, for example, auxiliary technological equipment: manipulators on the machine for setting a workpiece and removing a part in a flexible production module, maintenance of shelves in an automated warehouse, as well as low-precision machining, difficult profile welding, etc. In these cases, high precision drive mechanisms are no longer required. On the other hand, ball screws are not capable of moving the actuator more than 3000 mm. This forces us to look for more suitable mechanisms for the implementation of elementary linear movements in machines with parallel kinematics.

This article suggests using a rigid chain system. The most significant results in the creation of such chains have been achieved by Serapid and Tsubaki. Manufacturers indicate the following as the main advantages of their product [10, 11]:

- ease of installation and maintenance;
- high efficiency (up to 80%);
- accuracy, rigidity, repeatability;
- resistance to temperatures, humidity, aggressive environments, pollution;
- high carrying capacity (up to several hundred tons);
- high speed and acceleration;
- the ability to perform small movements with a low vibration level;
- low noise level;
- environmental safety and energy efficiency;
- compactness;
- ease of synchronization.

These companies position the rigid chain as a compact alternative to pneumatic and hydraulic cylinders when used in lifting and transport devices. At the same time, there are various modifications for various tasks (for horizontal and vertical movements, high-load solutions and non-metallic mechanisms such as a hard belt)

This transmission is based on the principle of a chain with a limited articulation. A roller chain is used which meshes with gears mounted on the drive shaft of the mechanical drive. The links are pivotally connected in such a way that they deviate relative to the main guide line in only one direction. When moving through the guide vanes, the links turn, the chain is rigidly fixed along a straight line, effectively working in tension and compression. At the same time, the chain is able to perceive bending loads in only one direction.

To ensure stiffness in bending in both directions, two chains are used at once, which mesh with each other (figure 1). In contrast to the "lead screw-nut" mechanism, in which the non-working part of the screw extends beyond the dimensions of the working area, the rigid chain can be folded, which makes it possible to organize its compact storage when rolled up.

Due to the large number of joints, the chain has a cumulative error. The maximum positioning accuracy in the best samples today does not exceed 1 mm with a chain length of 5000 mm.

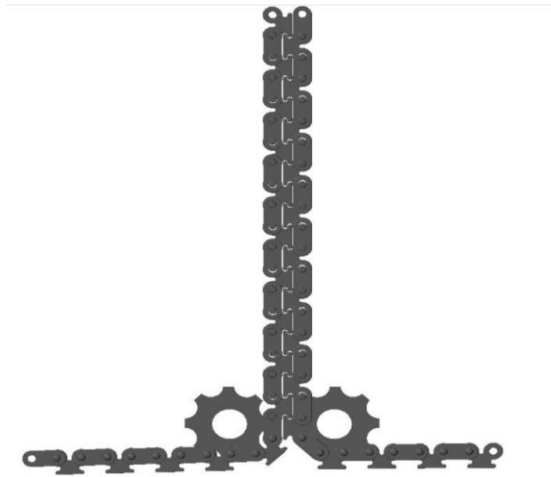

**Figure 1.** Rigid chain, principle of operation.

The production necessity of creating a manipulation robot capable of moving the actuator along 3 coordinate axes with the provision of a working area of at least 60% of the total volume of the production room was considered. Figure 2 shows a schematic solution to the problem. The main element of positioning is a universal floating platform, on which it is possible to install the required equipment, such as: a vision system, manipulation grips, special actuators. The platform is secured by three rigid chains. Each end of the chain is attached to the platform by means of a universal joint, allowing it to rotate freely along two axes (figure 3). The use of a gimbal is due to the angular movement between the rigid chain and the platform during the movement. Folded chain drives and storage housings are located in a triangle shape at the highest points of the industrial building.

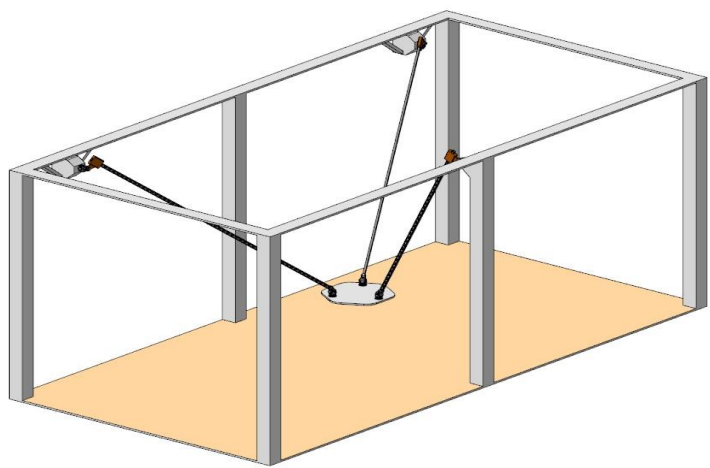

**Figure 2.** Scheme of a production room with a rigid chain positioning system.

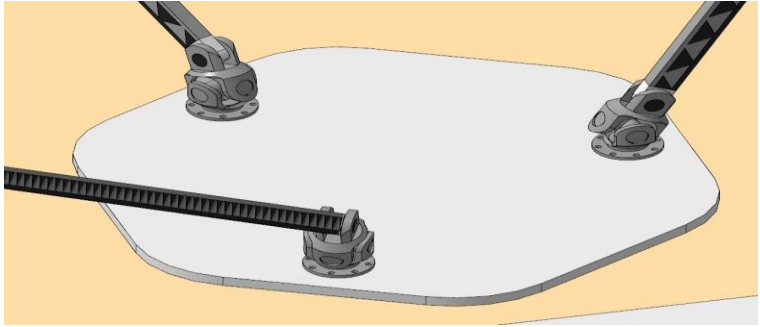

**Figure 3.** Gimbal platform suspension.

Moving the platform with the actuator is carried out by changing the lengths of three rigid chains by winding and unwinding them.

The main advantage of such a positional device is the high utilization of space, due to the location of the chain drive mechanisms almost close to the walls of the production room, which would be impossible with ball screws. In addition, this coefficient can be further increased if the platform is made with four rigid chains, the driving mechanisms of which are located in the corners of the production room. The use of a kinematically excessive amount of rigid chains allows increasing the bearing capacity of such a robot and arranging its structure more flexibly, taking into account the configuration of the production room.

## 3. Conclusion
A comparative analysis of the actuator positioning systems based on parallel kinematics is carried out. The use of kinematic systems using ball screws has a high positioning accuracy but does not provide system scalability. While the use of a rigid chain allows you to create a compact, rigid and scalable system. A method of using parallel kinematics to create a positioning system in an industrial room to ensure transportation and placement of a product during production is proposed.

**Acknowledgments**
This work was supported by the Ministry of Science and Higher Education of the Russian Federation (State Contract No. FEFE-2020-0017).